\documentclass[letterpaper]{article} 
\usepackage{aaai25}  
\usepackage{times}  
\usepackage{helvet}  
\usepackage{courier}  
\usepackage[hyphens]{url}  
\usepackage{graphicx} 
\usepackage{natbib}  
\usepackage{caption} 
\usepackage{algorithm}
\usepackage{algorithmic}
\usepackage{caption}
\usepackage{amsmath,amsfonts}
\usepackage{array}
\usepackage{colortbl}
\usepackage{textcomp}
\usepackage{url}
\usepackage{booktabs}
\usepackage{verbatim}
\usepackage{graphicx}
\usepackage{cite}
\usepackage{tabularx}
\usepackage{color}
\usepackage{xcolor}
\usepackage{multirow}
\usepackage{colortbl}
\usepackage{subfigure}
\newcommand{\vcenteredbox}[1]{\begingroup
\setbox0=\hbox{#1}\parbox{\wd0}{\box0}\endgroup}

\usepackage{newfloat}
\usepackage{listings}
\DeclareCaptionStyle{ruled}{labelfont=normalfont,labelsep=colon,strut=off} 
\floatstyle{ruled}
\newfloat{listing}{tb}{lst}{}
\floatname{listing}{Listing}
\title{VLScene: Vision-Language Guidance Distillation for Camera-Based \\ 3D Semantic Scene Completion}
\author{
    Meng Wang,
    Huilong Pi\equalcontrib,
    Ruihui Li,
    Yunchuan Qin,
    Zhuo Tang\equalcontrib,
    Kenli Li
}
\affiliations{
    College of Computer Science and Electronic Engineering, Hunan University, Changsha, China\\

    \{willem, liruihui, phl880217, qinyunchuan, ztang, lkl\}@hnu.edu.cn
}

\usepackage{bibentry}

\begin{document}

\maketitle
\begin{abstract}
Camera-based 3D semantic scene completion (SSC) provides dense geometric and semantic perception for autonomous driving. However, images provide limited information making the model susceptible to geometric ambiguity caused by occlusion and perspective distortion. Existing methods often lack explicit semantic modeling between objects, limiting their perception of 3D semantic context.
To address these challenges, we propose a novel method VLScene: Vision-Language Guidance Distillation for Camera-based 3D Semantic Scene Completion. The key insight is to use the vision-language model to introduce high-level semantic priors to provide the object spatial context required for 3D scene understanding.
Specifically, we design a vision-language guidance distillation process to enhance image features, which can effectively capture semantic knowledge from the surrounding environment and improve spatial context reasoning. In addition, we introduce a geometric-semantic sparse awareness mechanism to propagate geometric structures in the neighborhood and enhance semantic information through contextual sparse interactions.
Experimental results demonstrate that VLScene achieves rank-1st performance on challenging benchmarks—SemanticKITTI and SSCBench-KITTI-360, yielding remarkably mIoU scores of 17.52 and 19.10, respectively.

\end{abstract}
\begin{links}
    \link{Code}{https://github.com/willemeng/VLScene}
\end{links}
\section{Introduction}
\label{sec:intro}

The field of 3D perception faces new challenges with the emergence of autonomous driving. 
Autonomous vehicles need to predict the surrounding environment accurately to ensure safe navigation and obstacle avoidance. 
The Semantic Scene Completion (SSC)~\cite{song2017semantic,roldao2020lmscnet,yan2021sparse} task aims to forecast the semantic occupancy of every voxel in the entire 3D scene based on limited observations.

Most existing SSC~\cite{rist2021semantic,zhang2018efficient,guo2018view,li2020aicnet,roldao2020lmscnet,yan2021sparse} solutions rely on input RGB images and corresponding 3D data to predict volume occupancy and semantic labels. However, the reliance on 3D data often necessitates the use of specialized and expensive depth sensors, which limits the broader application of SSC algorithms. Recently, many researchers~\cite{cao2022monoscene,li2023voxformer,li2023stereoscene,jiang2024symphonize,wang2024h2gformer,xue2024bi} have explored the use of camera-based approaches to recover dense 3D geometric structures and semantic information.

\begin{figure}[t]
  \centering
  \includegraphics[width=0.98\linewidth]{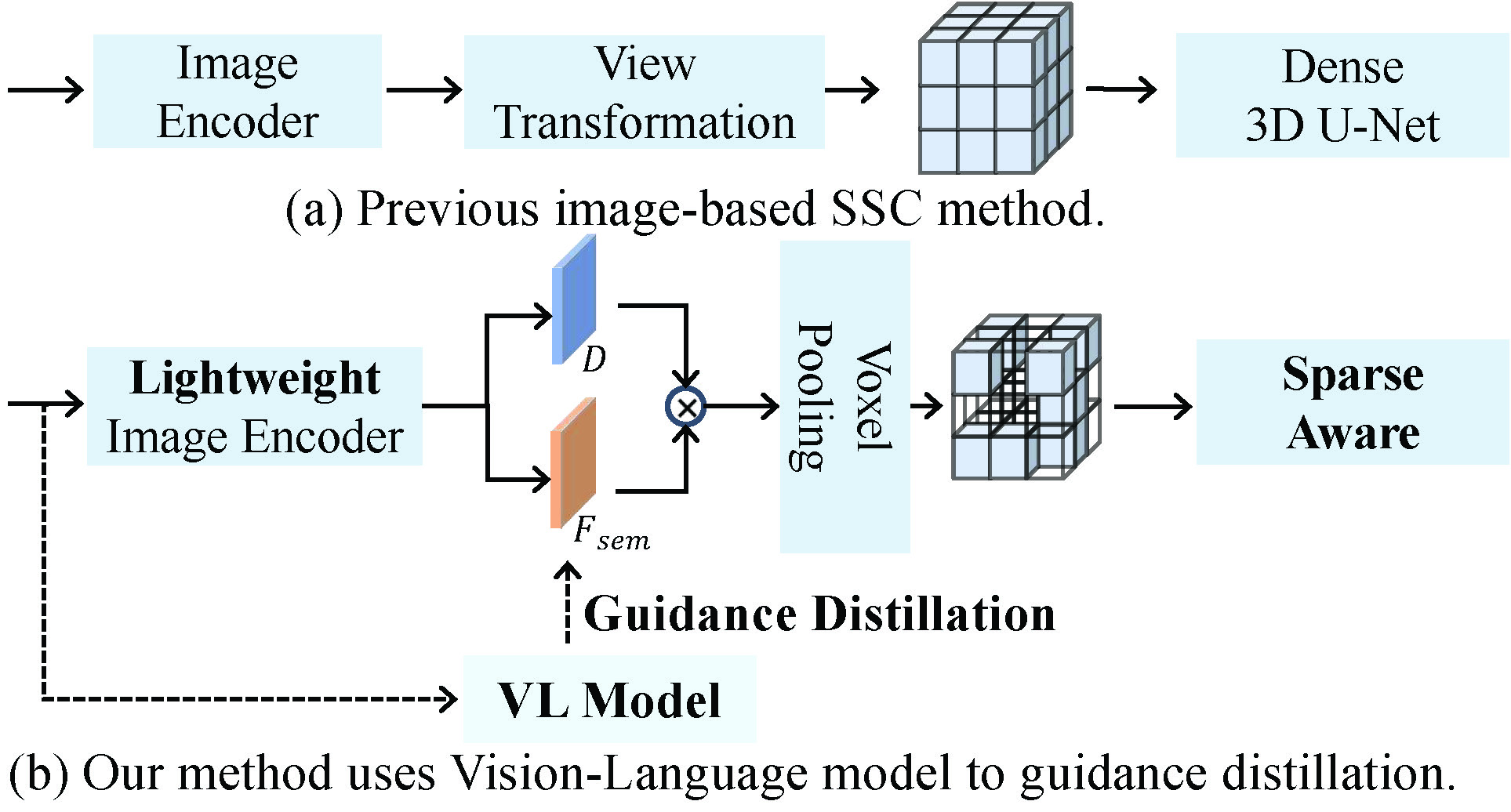}
  \caption{Our method uses vision-language model guidance distillation versus the previous image-based method.}
  \label{fig:figure1}
\end{figure}

In Figure~\ref{fig:figure1}, existing methods commonly depend on 2D-3D view transformations to construct 3D representations from image features, and utilise complex 3D models for geometric and semantic inference. Nevertheless, the limited information available from images renders these models susceptible to geometric ambiguities caused by occlusions and perspective distortions. Inferring geometry without sufficient visual input requires leveraging semantic knowledge of the surrounding environment and reasoning about the spatial context.
Given these limitations, we consider the question: \textbf{\textit{How can high-level semantic priors be utilized to improve semantic representation and spatial context?}}

\begin{figure}[t]   
\centering
\quad
\subfigure[]{
\includegraphics[width=0.81\linewidth]{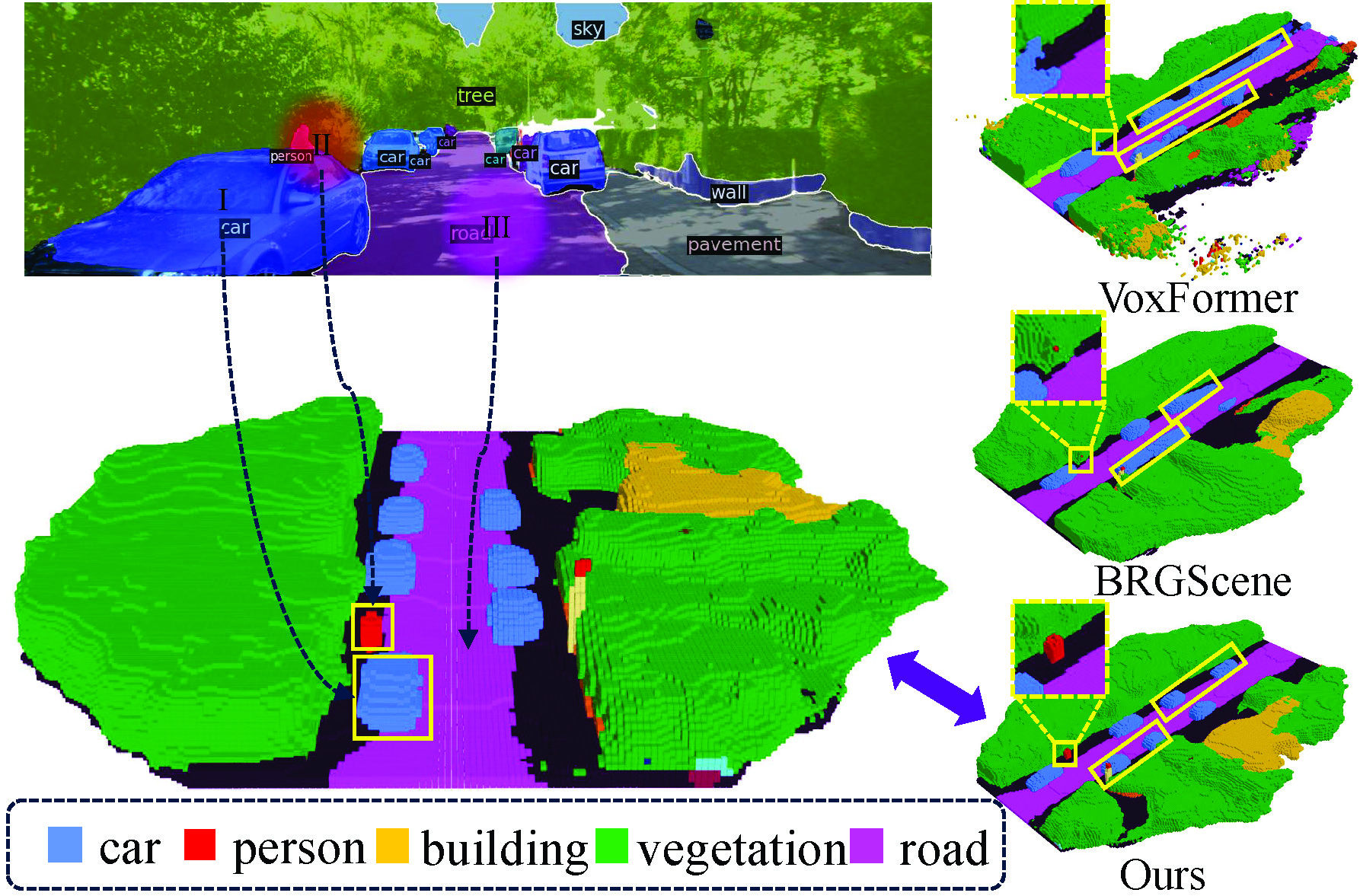}
}\\
\quad
\subfigure[]{
\includegraphics[width=0.81\linewidth]{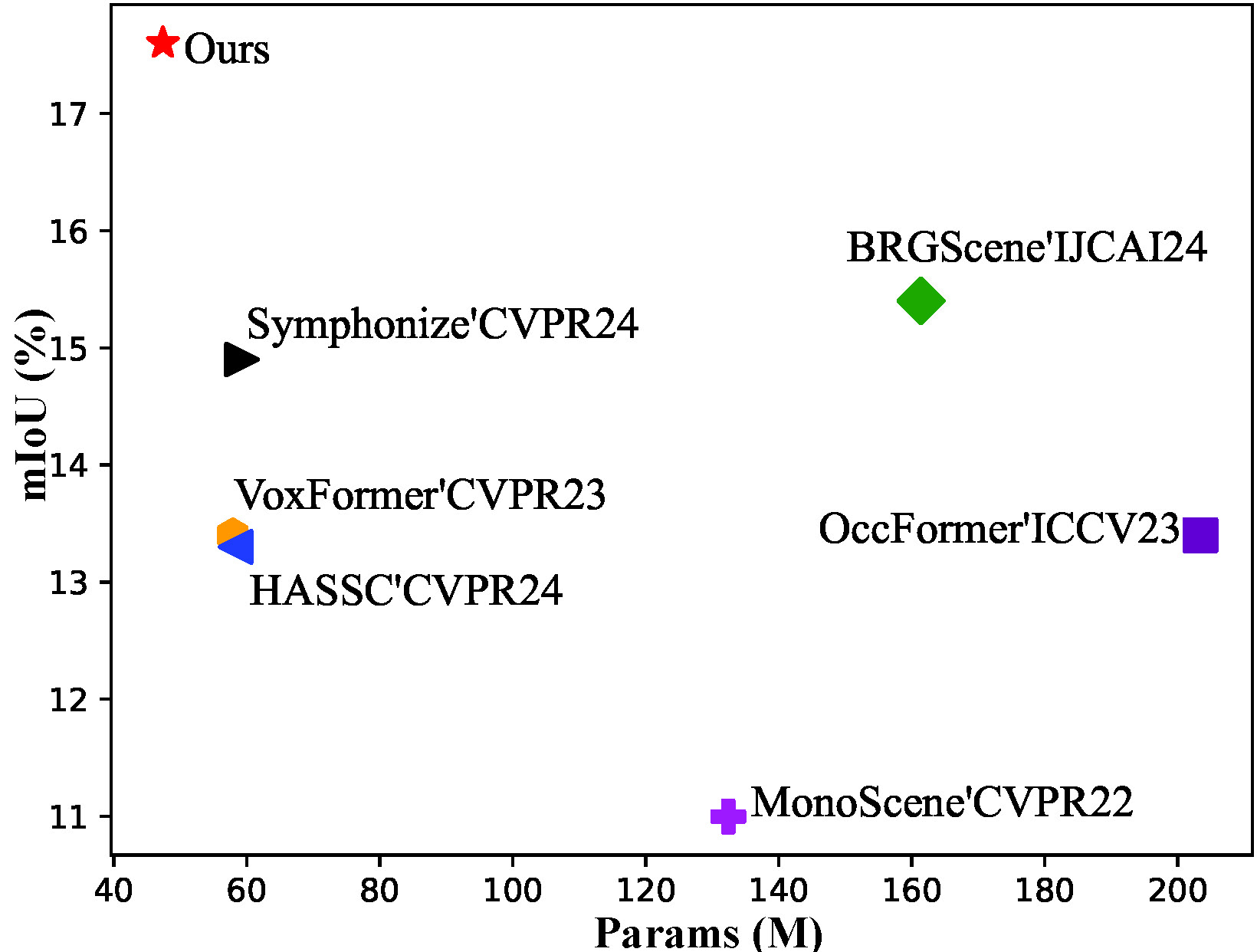}
}
\caption{(a) VLScene uses 2D high-level semantic priors improve 3D scene. (b) Comparison of Params and mIoU.}
\label{fig:figure2}
\end{figure}

In this paper, we propose a novel camera-based SSC method: VLScene, Vision-Language Guidance Distillation for Camera-based 3D Semantic Scene Completion. VLScene leverages a vision-language model to extract high-level semantic priors, enhancing semantic representation and spatial context through distillation.
As shown in the yellow box in Figure~\ref{fig:figure2}(a), semantic and spatial structure priors are obtained through vision-language models. The position and category of the person behind the car are accurately inferred, and even densely packed cars in the image are effectively distinguished in the corresponding scene.
Specifically, we introduce \textit{vision-language guidance distillation} to improve image features, which effectively captures the semantic knowledge of the surrounding environment and enhances reasoning about the spatial context. We further design a \textit{geometric-semantic sparse awareness} mechanism, consisting of two modules: neighborhood geometry propagation (NGP) and sparse semantic interaction (SSI), to sparsely perceive voxel information from both geometric and semantic perspectives.
The NGP module alternates between large and small kernel convolutions to ensure comprehensive capture of objects of varying sizes in the adapted 3D scene. Meanwhile, SSI employs sparse convolutions to effectively utilize the geometric information in voxel features and enhances semantic information through contextual interactions.
To evaluate the performance of VLScene, we conduct thorough experiments on SemanticKITTI~\cite{behley2019semantickitti} and SSCBench-KITTI360~\cite{Liao2022kitti360,li2023sscbench}. As shown in Figure~\ref{fig:figure2}(b), our method achieves state-of-the-art performance with a mIoU of 17.52\%, using only 47.4M parameters.

\begin{itemize}
    \item  We propose a novel method, VLScene, leveraging vision-language models to extract high-level semantic priors, thereby enhancing semantic representation and spatial context through a distillation process.
    \item We design vision-language guidance distillation that effectively captures semantic knowledge of the surrounding environment and improves spatial context reasoning.
    \item We introduce geometric-semantic sparse awareness to propagate geometric structure in the neighborhood and enhance semantic information through contextual sparse interactions.
    \item The proposed VLScene model achieves SOTA results on SemanticKITTI and SSCBench-KITTI-360 benchmarks, surpassing the latest methods.

\end{itemize}

\section{Related Work}
\label{sec:rel}

\subsection{3D Semantic Scene Completion}
The SSC task was designed to address the challenges of scene completion and semantic segmentation by predicting the occupancy and semantic categories of each voxel in a 3D scene. SSCNet~\cite{song2017semantic} was the first to define the semantic scene completion task, wherein both geometry and semantics were jointly inferred from incomplete visual observations. Subsequent research recognized the inherently 3D nature of this task, leading to numerous studies that directly employed 3D inputs~\cite{rist2021semantic,zhang2018efficient,Xia_2023_scpnet}, such as depth information, occupancy meshes, and point clouds, to leverage their rich geometric cues. To incorporate additional texture information, many works~\cite{cai2021semantic,li2019rgbd} explored the use of multi-modal inputs, combining RGB images with diverse geometric cues.
As purely visual autonomous driving solutions became more cost-effective, MonoScene~\cite{cao2022monoscene} was the first to infer dense geometry and semantics from a single monocular RGB image. TPVFormer~\cite{huang2023tri} proposed a three-perspective view representation, which, along with BEV, introduced two additional vertical planes. OccFormer~\cite{zhang2023occformer} employed a dual-path transformer framework to encode 3D voxel features. VoxFormer~\cite{li2023voxformer} adopted a novel two-stage framework to elevate images into fully 3D voxelized semantic scenes.
NDCScene~\cite{yao2023ndc} extended the 2D feature map to a normalized device coordinate space instead of the world space. MonoOcc~\cite{zheng2024monoocc} further enhanced the 3D volume with an image-conditioned cross-attention module. H2GFormer~\cite{wang2024h2gformer} effectively utilized 2D features through a progressive feature reconstruction process across various directions. Symphonize~\cite{jiang2024symphonize} extracted high-level instance features from the image feature map, serving as the key and value for cross-attention. HASSC~\cite{wang2024HASSC} introduced a self-distillation training strategy to improve the performance of VoxFormer. Finally, BRGScene~\cite{li2023stereoscene} utilized binocular image inputs to implicitly generate stereo depth information and employed stereo matching to resolve geometric ambiguities.

\begin{figure*}[ht]
\centering
  \includegraphics[width=\textwidth]{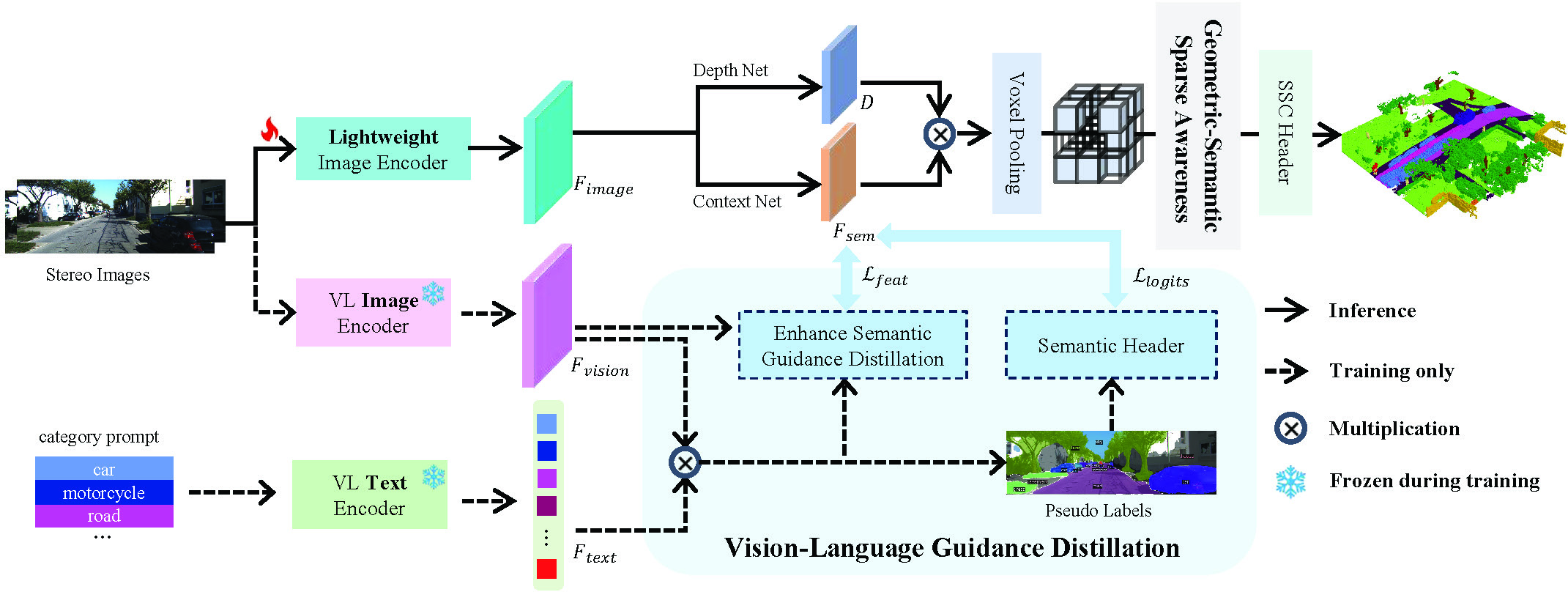}
  \caption{The VLScene framework is proposed for camera-based 3D semantic scene completion. }
  \label{fig:figure3}
\end{figure*}

\subsection{Large Vision-Language Model}

Large vision-language models, such as Contrastive Vision-Language Pre-training (CLIP)~\cite{radford2021clip}, achieved significant advancements in open-set and zero-shot image classification tasks due to their strong alignment between visual and text embeddings. LSeg~\cite{li2022lseg} focused on aligning pixel-level image features with class embeddings generated by the CLIP text encoder. MaskCLIP~\cite{dong2023maskclip} explored dense prediction problems using CLIP by making minor adjustments to the network structure.
However, practical applications like autonomous driving and indoor navigation required a deeper understanding of 3D scenes. Consequently, recent research investigated the application of 2D vision-language pre-training for 3D perception tasks. For instance, CLIP2Scene~\cite{chen2023clip2scene} introduced a semantically driven cross-modal contrastive learning framework. OpenScene~\cite{Peng2023OpenScene} used a pre-trained VL model~\cite{ghiasi2022openseg, kuo2023F-VLM} to extract per-pixel CLIP features and projected 3D points onto the image plane to derive dense 3D features. RegionPLC~\cite{yang2023regionplc} leveraged regional visual cues to generate dense captions and performed point-discriminative contrastive learning.
\subsection{Cross-modal Knowledge Distillation}
The primary objective of cross-modal knowledge distillation was to transfer knowledge between different modalities. Knowledge distillation~\cite{hinton2015distilling,wang2022head,mirzadeh2020improved} was initially introduced for model compression, focusing on transferring the learned knowledge from a teacher network to a student network. With advancements in multi-sensor technologies, 2D-to-3D distillation methods were developed to enable models to utilize data from various modalities, thereby enhancing their performance in 3D tasks. For example, PPKT~\cite{liu2021PPKT} employed InfoNCE loss to assist 3D networks in extracting valuable knowledge from 2D image backbones.
2DPass~\cite{yan20222dpass} proposed an innovative approach to improve semantic information extraction from multimodal data by integrating auxiliary modality fusion and multi-scale fusion into a single distillation framework. CMKD~\cite{hong2022cmkd} effectively transferred point cloud features and responses to images, significantly boosting performance. BEVDistill~\cite{chen2022bevdistill} unified the two modalities into BEV space for distillation, achieving more accurate object detection by using a grouping network to form a global descriptor. UniDistill~\cite{zhou2023unidistill} focused on BEV object detection and leveraged knowledge distillation in features, relations, and responses.
\section{Methodology}
\paragraph{Problem setup.}
Given a set of stereo RGB images $I_{stereo} = \{I_{l},I_{r}\}$. The goal is to infer the geometry and semantics of the 3D scene jointly. The scene is represented as a voxel grid ${Y}\in\mathbb{R}^{X \times Y \times Z \times{(M+1)} }$, where X, Y, Z represent height, width and depth in 3D space. For each voxel, it will be assigned to a unique semantic label belonging to $C\in\{C_0, C_1, ..., C_M\}$ that either occupies the empty space $C_0$ or falls in a specific semantic class $\{C_1, ..., C_M\}$. Here M represents the total number of semantic classes. We want to learn a transformation $Y=\theta(I_{stereo})$ as close to the ground truth $\hat{Y}$ as possible.


\paragraph{Overview.}
We illustrate our approach in Figure~\ref{fig:figure3}. We first use the lightweight image encoder RepViT~\cite{wang2023repvit} and FPN~\cite{lin2017feature} to extract image features $F_{image}$, and use the vision-language model LSeg~\cite{li2022lseg} to extract visual features $F_{vision}$ and text features $F_{text}$. $F_{image}$ passes through the deep network and the context network to obtain discrete depth values $D$ and semantic features $F_{sem}$.
Subsequently, we use the VLGD module to improve the semantic features with the high-level semantic information captured by the VL model, and then obtain the voxel features $V$ through the LSS view transformation. Then, $V$ enters the GSSA module to sparsely perceive the voxel information from the geometric and semantic perspectives, and obtains the refined voxel features $V_{fine}$. Finally, $V_{fine}$ outputs dense semantic voxels $Y$ through upsampling and linear projection.

\subsection{Vision-Language Guidance Distillation}
\label{sec:vlgd}
Existing camera-based SSC methods often lack explicit semantic modeling and are susceptible to geometric ambiguities caused by occlusion and perspective distortion. To enrich semantic representations, we introduce a vision-language guidance distillation module in 2D space. This module leverages the VL model to extract high-level semantic priors, enhancing both semantic representations and spatial context through distillation. Details are shown in Figure~\ref{fig:figure3}.
\begin{figure}[t]
  \centering
  \includegraphics[width=\linewidth]{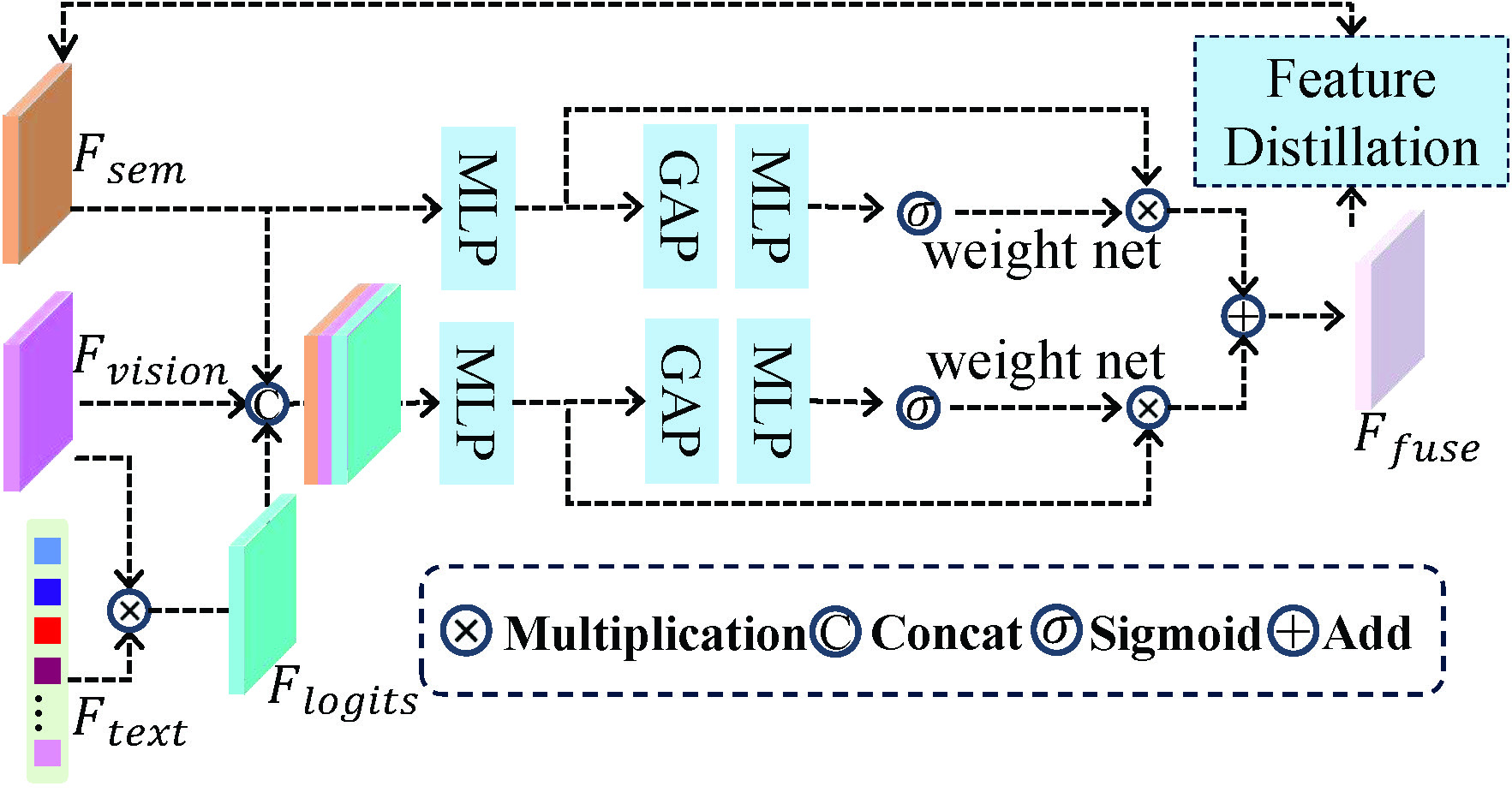}
  \caption{Illustration of the enhanced semantic feature distillation.}
  \label{fig:figure4}
\end{figure}
\paragraph{Feature Distillation.}
Given an input image $I_l$, we extract image features $F_{image}$ using a lightweight image encoder and visual features $F_{vision}$ using a VL image encoder. Simultaneously, we input the category name as a prompt into the VL text encoder to obtain text features $F_{text} \in \mathbb{R}^{Q\times C}$, where $Q$ represents the number of categories and $C$ denotes the feature channels. We then compute the cosine similarity between the VL image and text features to generate a 2D semantic map:
\begin{equation}
F_{logits}=\underset{q \in\{1, \ldots, Q\}}{\text{softmax} } \frac{F_{{vision }} \otimes F_{text}^{\top}}{\left\|F_{{vision }}\right\|\|F_{text}\|},
\end{equation}
where $\otimes$ denotes matrix multiplication, and $F_{logits}$ denotes the segmentation map by maximizing the similarity scores between VL image and text features along the category dimension.

To effectively select rich semantic cues from the VL model, we adaptively fuse the semantic information from both the image encoder and the VL model to obtain enhanced fused semantic features, which are then subjected to feature distillation. Figure~\ref{fig:figure4} illustrates the detailed process of this enhanced semantic feature distillation. First, we concatenate the three features $F_{sem}$, $F_{vision}$, and $F_{logits}$ and perform an initial fusion using two convolutional layers to obtain the feature $\hat{F}_{vision}$. Next, we calculate the channel attention for $\hat{F}_{vision}$ and $F_{sem}$ to adaptively weight the feature channels,
\begin{equation}
\begin{split}
    \hat{F}_{vision}^{weight} =\mathbf{\sigma}[\text{GAP}(\text{MLP}(\hat{F}_{vision}))],\\
    F_{sem}^{weight} =\mathbf{\sigma}[\text{GAP}(\text{MLP}(F_{sem}))],\\
\end{split}
\end{equation}
where $\mathbf{\sigma}$ denotes the sigmoid function, GAP represents the global average pooling operation. The weighted features are then summed to obtain the fused feature $F_{fuse}$, 
\begin{equation}
\begin{split}
    F_{fuse} = \hat{F}_{vision}^{weight}*\text{MLP}(\hat{F}_{vision}) \\
            +F_{sem}^{weight}*\text{MLP}(F_{sem}).
\end{split}
\end{equation}
For each pixel feature, we calculate the difference between its features on $F_{sem}$ and $F_{fuse}$ the feature distillation loss $\mathcal{L}_{kd{\_}feat}$:
\begin{equation}
\mathcal{L}_{kd{\_}feat} = \| F_{sem} - F_{fuse}  \|_1.
\end{equation}
\paragraph{Logits Distillation.} We also perform explicit logit distillation on the semantic features. Specifically, we pass $F_{sem}$ through the image semantic head composed of residual blocks to obtain the image segmentation result $F_{pred}$ of category Q. Then calculate the loss with the image semantic pseudo label $F_{logits}$ obtained by the vision-language model. The specific operation is as follows:
\begin{equation}
    \mathcal{L}_{kd{\_}logits} = \operatorname{CrossEntropyLoss}(F_{pred},F_{logits}),
\end{equation}
After our designed vision-language guidance distillation, we will get the enhanced semantic feature $\hat{F}_{sem}$. Then, we follow the view transformation module of BRGScene~\cite{li2023stereoscene} and construct the voxel feature V using $\hat{F}_{sem}$.

\begin{figure}[t]
  \centering
  \includegraphics[width=\linewidth]{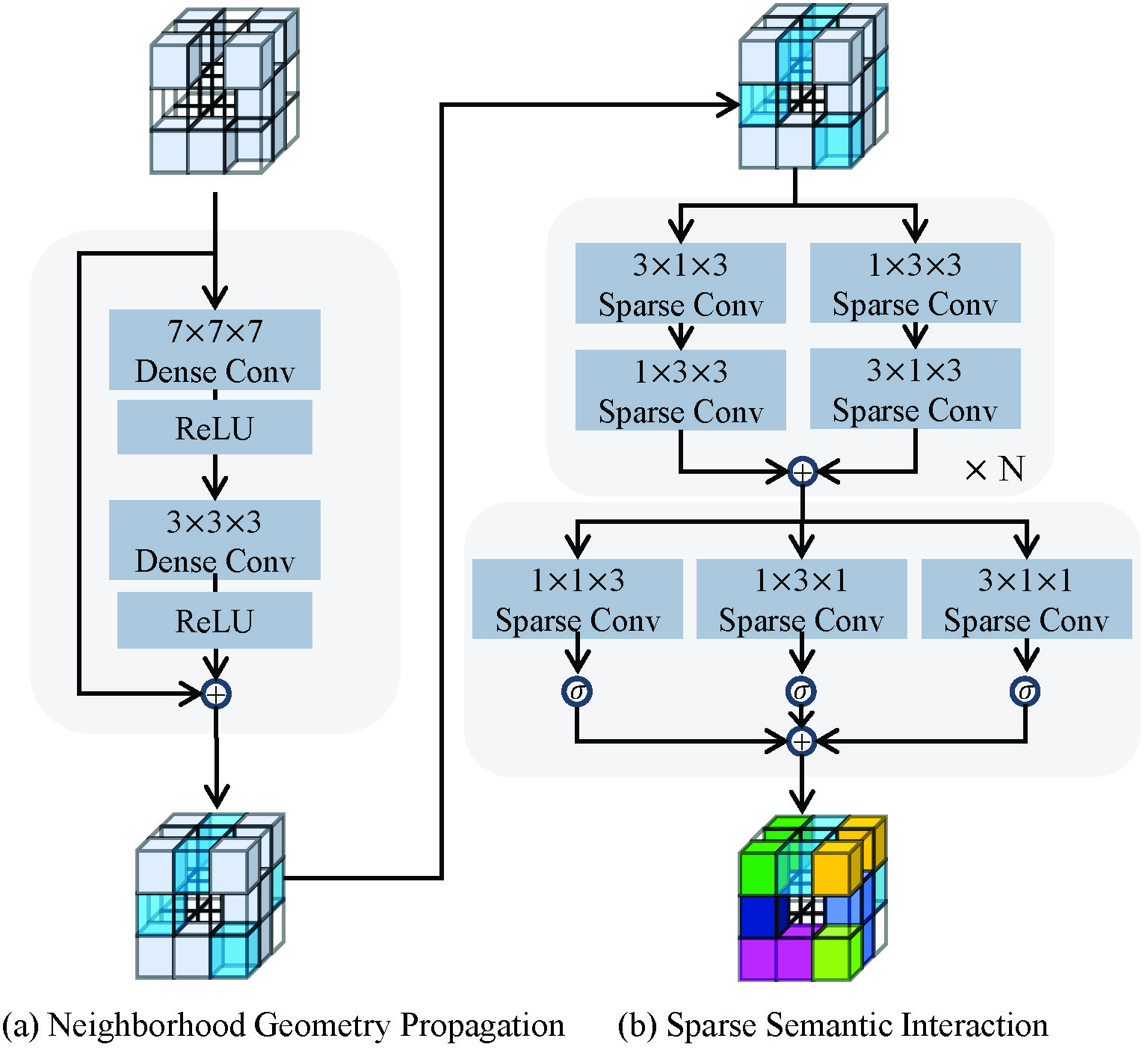}
  \caption{Illustration of the geometric-semantic sparse awareness.}
  \label{fig:figure5}
\end{figure}

\begin{figure*}[ht]
\centering
  \includegraphics[width=0.95\textwidth]{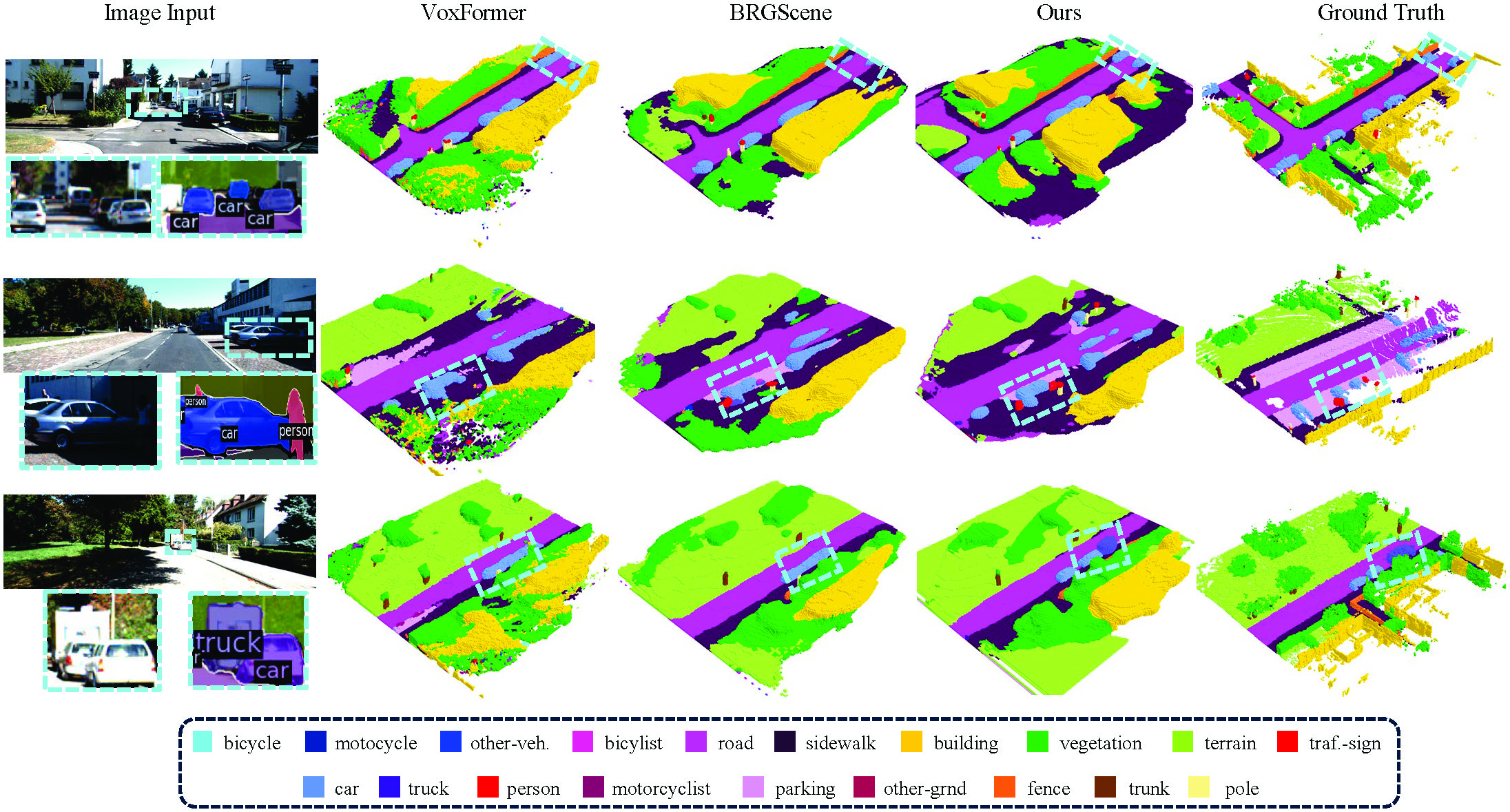}
  \caption{Qualitative results on the SemanticKITTI validation set.}
  \label{fig:figure6}
\end{figure*}

\subsection{Geometric-Semantic Sparse Awareness}
\label{sec:gssa}
It is observed that even after projecting the enhanced semantic features into voxel space, approximately half of the voxels remain empty (especially out-of-field, obstructed and distant areas). To fully utilize the available voxel features, we introduce the geometric-semantic sparse awareness module, which can perceive voxel information both geometrically and semantically. The detailed process of geometric-semantic sparse awareness is illustrated in Figure~\ref{fig:figure5}.

\paragraph{Neighborhood Geometry Propagation.}
Geometric information is typically extracted only from existing non-empty voxels, leading to a lack of features in empty voxels. To address this, we design the Neighborhood Geometry Propagation (NGP) module, which diffuses voxel features into adjacent empty regions. The detailed process is shown in Figure~\ref{fig:figure5}(a).

\begin{table*}[t]
  \centering
  \small
  \setlength{\tabcolsep}{2pt}
  \begin{tabular}{l|c|ccccccccccccccccccc|c}
    \toprule
    \textbf{Methods} &\textbf{IoU}   
    & \rotatebox{90}{\textbf{road}} 
    & \rotatebox{90}{\textbf{sidewalk}} 
    & \rotatebox{90}{\textbf{parking}} 
    & \rotatebox{90}{\textbf{other-grnd}} 
    &  \rotatebox{90}{\textbf{building}} 
    &  \rotatebox{90}{\textbf{car}} 
    & \rotatebox{90}{\textbf{truck}} 
    & \rotatebox{90}{\textbf{bicycle}}  
    & \rotatebox{90}{\textbf{motocycle}} 
    & \rotatebox{90}{\textbf{other-vehicle}} 
    & \rotatebox{90}{\textbf{vegetation}} 
    & \rotatebox{90}{\textbf{trunk}} 
    & \rotatebox{90}{\textbf{terrain}} 
    & \rotatebox{90}{\textbf{person}} 
    & \rotatebox{90}{\textbf{bicylist}} 
    & \rotatebox{90}{\textbf{motorcyclist}} 
    &  \rotatebox{90}{\textbf{fence}} 
    & \rotatebox{90}{\textbf{pole}} 
    & \rotatebox{90}{\textbf{traf.-sign}}& \textbf{mIoU}  \\
    
    \midrule

    MonoScene & 34.16  & 54.7&27.1&24.8&5.7&14.4&18.8&3.3&0.5&0.7&4.4&14.9&2.4&19.5&1.0&1.4&0.4&11.1&3.3&2.1 & 11.08 \\
    
    TPVFormer& 34.25 &55.1&27.2&27.4&6.5&14.8&19.2&3.7&1.0&0.5&2.3&13.9&2.6&20.4&1.1&2.4&0.3&11.0&2.9&1.5& 11.26\\

    OccFormer& 34.53  & 55.9&30.3&31.5&6.5&15.7&21.6&1.2&1.5&1.7&3.2&16.8&3.9&21.3&2.2&1.1&0.2&11.9&3.8&3.7& 12.32\\ 
    VoxFormer& 43.21& 54.1& 26.9& 25.1& 7.3& 23.5& 21.7& 3.6& 1.9& 1.6& 4.1& 24.4& 8.1& 24.2& 1.6& 1.1& 0.0& 13.1& 6.6& 5.7& 13.41\\

    MonoOcc&-&55.2&27.8&25.1&9.7&21.4&23.2&5.2&2.2&1.5&5.4&24.0&8.7&23.0&1.7&2.0&0.2&13.4&5.8&6.4&13.80\\
    H2GFormer&44.20&56.4&28.6&26.5&4.9&22.8&23.4&4.8&0.8&0.9&4.1&24.6&9.1&23.8&1.2&2.5&0.1&13.3&6.4&6.3&13.72\\
    HASSC&43.40&54.6&27.7&23.8&6.2&21.1&22.8&4.7&1.6&1.0&3.9&23.8&8.5&23.3&1.6&4.0&0.3&13.1&5.8&5.5&13.34\\
    Symphonize&42.19&58.4&29.3&26.9&11.7&24.7&23.6&3.2&3.6&2.6&5.6&24.2&\textbf{10.0}&23.1&\textbf{3.2}&1.9&\textbf{2.0}&16.1&7.7&8.0&15.04\\
    BRGScene & 43.34 & 61.9  &31.2  &30.7 & 10.7 & 24.2 & 22.8 & 2.8 & 3.4 & 2.4 & 6.1 & 23.8 & 8.4 & 27.0 & 2.9 & 2.2 & 0.5 & 16.5 & 7.0 & 7.2 & 15.36\\
    \rowcolor{gray!20}\textbf{Ours}  & \textbf{45.14} &\textbf{64.7}&\textbf{34.7}&\textbf{32.4}&\textbf{13.1}&\textbf{27.3}&\textbf{26.1}&\textbf{6.5}&\textbf{4.2}&\textbf{3.8}&\textbf{8.3}&\textbf{26.4}&\textbf{10.0}&\textbf{29.4}&2.8&\textbf{5.1}&0.9&\textbf{20.0}&\textbf{8.9}&\textbf{8.4}& \textbf{17.52}\\
    
    
    



    \bottomrule
  \end{tabular}
    \caption{Quantitative results on the SemanticKITTI hidden test set. {\textbf{Bold}} denotes the best performance.}
  \label{tab:SemanticKITTI Quantitative Comparison}
\end{table*}
\begin{table*}[t]
  \centering
  \small
  \setlength{\tabcolsep}{2pt}

  \begin{tabular}{l|c|cccccccccccccccccc|c}
    \toprule
    \textbf{Methods} &\textbf{IoU}   
    &  \rotatebox{90}{\textbf{car} } 
    & \rotatebox{90}{\textbf{bicycle}}  
    & \rotatebox{90}{\textbf{motocycle}} 
    & \rotatebox{90}{\textbf{truck}} 
    & \rotatebox{90}{\textbf{other-vehicle}} 
    & \rotatebox{90}{\textbf{person}} 
    &  \rotatebox{90}{\textbf{road}} 
    & \rotatebox{90}{\textbf{parking}} 
    & \rotatebox{90}{\textbf{sidewalk}} 
    & \rotatebox{90}{\textbf{other-grnd}} 
    & \rotatebox{90}{\textbf{building}} 
    &  \rotatebox{90}{\textbf{fence}} 
    & \rotatebox{90}{\textbf{vegetation}} 
    & \rotatebox{90}{\textbf{terrain}} 
    & \rotatebox{90}{\textbf{pole}} 
    & \rotatebox{90}{\textbf{traf.-sign}} 
    & \rotatebox{90}{\textbf{other-struct.}} 
    & \rotatebox{90}{\textbf{other-obj.}} &\textbf{mIoU}   \\
    
    \midrule
    MonoScene&37.87&19.3&0.4&0.6&8.0&2.0&0.9&48.4&11.4&28.1&3.3&32.9&3.5&26.2&16.8&6.9&5.7&4.2&3.1&12.31\\
    VoxFormer&38.76&17.8&1.2&0.9&4.6&2.1&1.6&47.0&9.7&27.2&2.9&31.2&5.0&29.0&14.7&6.5&6.9&3.8&2.4&11.91\\
    TPVFormer&40.22&21.6&1.1&1.4&8.1&2.6&2.4&53.0&12.0&31.1& 3.8&34.8&4.8&30.1&17.5&7.5&5.9&5.5&2.7&13.64\\
    OccFormer&40.27&22.6&0.7&0.3&9.9&3.8&2.8&54.3&13.4&31.5&3.6&36.4&4.8&31.0&19.5&7.8&8.5& 7.0&4.6&13.81\\
    Symphonies&44.12&\textbf{30.0}&1.9&5.9&\textbf{25.1}&\textbf{12.1}&\textbf{8.2}&54.9&13.8&32.8&\textbf{6.9}&35.1&8.6&\textbf{38.3}&11.5&14.0&9.6&\textbf{14.4}&\textbf{11.3}&18.58\\
    \rowcolor{gray!20}\textbf{Ours}& \textbf{46.08} &29.0&\textbf{4.7}&\textbf{7.7}&18.3&7.6&7.4&\textbf{60.1}&\textbf{17.4}&\textbf{39.0}&6.0&\textbf{42.1}&\textbf{9.6}&36.5&\textbf{24.8}&\textbf{17.0}&\textbf{18.8}&10.5&6.5& \textbf{19.10}\\
    \bottomrule
  \end{tabular}
    \caption{Quantitative results on the SSCBench-KITTI360 test set. {\textbf{Blod}} denotes the best performance.}
  \label{tab:KITTI360Test Quantitative Comparison}
\end{table*}

Specifically, NGP consists of alternating $7\times7\times7$ large kernel convolution layers and $3\times3\times3$ small kernel convolution layers. The large kernel convolutions handle large objects (such as cars and trucks) and view boundaries, propagating features from non-empty voxels to surrounding empty ones, thereby capturing the complete geometric structure of larger objects. Conversely, small kernel convolutions target smaller objects, such as poles and pedestrians, providing finer granularity in feature propagation to accurately capture the complex details of these smaller objects. Additionally, inspired by SCPNet~\cite{Xia_2023_scpnet}, we remove the convolution bias and batch normalization layers to reduce computational costs and maintain the efficiency of subsequent sparse convolutions.
The specific calculation process is as follows:
\begin{equation}
    V_{{com}} = V + \mathbf{\delta} \left( \text{Conv}_{3\times 3\times 3} \left( \mathbf{\delta} \left( \text{Conv}_{7\times 7\times 7}(V) \right) \right) \right),
\end{equation}
where $\mathbf{\delta}$ denotes the ReLU activation function. Through the above operation, we obtain the completion voxel feature.
\paragraph{Sparse Semantic Interaction.}
To enrich the semantic information of the scene, we introduce a Sparse Semantic Interaction (SSI) module. This module effectively leverages the geometric information within voxel features and enhances semantic content through contextual interactions.

Initially, the completion feature $V_{com}$ is converted into a sparse representation by isolating non-empty voxels. The sparse tensor is stored in a commonly used coordinate format:
\begin{equation}
    V_{com} = \{\text{P}=[x,y,z],\text{F}\in\mathbb{R}^{N\times C}\},
\end{equation}
where $N$ represents the number of non-empty voxels, and P and F denote the coordinates and features of the voxels.

Subsequently, inspired by the observations and conclusions in Cylinder3D~\cite{zhu2021cylindrical}, we decompose the traditional sparse convolution kernel into two orthogonal kernels oriented in vertical and horizontal directions. This decomposition aligns with the voxel distribution of objects in the driving scene, enabling the orthogonal kernels to capture specific semantic information along both directions. By configuring two parallel and asymmetric convolution kernels, we further enhance the semantic richness of the scene.
Finally, following Cylinder3D, we apply three Rank-1 sparse kernels to extract low-rank features, which are then aggregated to produce refined voxel features, $V_{fine}$.

\subsection{Training Loss}
In the VLScene framework, we adopt the scene-class affinity loss from MonoScene~\cite{cao2022monoscene} to optimize precision, recall, and specificity concurrently.
The semantic scene completion loss is shown as follows:
\begin{equation}
     \mathcal{L}_{ssc} = \mathcal{L}^{sem}_{scal} + \mathcal{L}^{geo}_{scal}+ \mathcal{L}_{ce}+\mathcal{L}_{depth},
\end{equation}
The knowledge distillation loss is expressed as:
\begin{equation}
    \mathcal{L}_{kd} = \mathcal{L}_{kd\_feat}+\mathcal{L}_{kd\_logits},
\end{equation}
The overall training loss function is formulated as follows: 
\begin{equation}
        \mathcal{L} = \lambda_{ssc}\mathcal{L}_{ssc} + \lambda_{kd}\mathcal{L}_{kd},
\end{equation}
where several $\lambda$ are balancing coefficients.

\section{Experiments}
\label{sec:exp}
To assess the effectiveness of our VLScene, we conducted thorough experiments using the large outdoor datasets SemanticKITTI~\cite{behley2019semantickitti} and SSCBench-KITTI-360~\cite{li2023sscbench}. 
\subsection{Qualitative Results}
To intuitively demonstrate VLScene performance, Figure~\ref{fig:figure6} presents the qualitative results of VoxFormer, BRGScene, and our method on the SemanticKITTI validation set. The enlarged area in the first column shows the semantic prior of the scene object structure.
Compared to VoxFormer and BRGScene, our method more completely and accurately reconstructs distant objects, such as the car in the blue box in the first row. Additionally, our VLScene effectively captures the position and details of small objects, like the person in the blue box in the second row. It demonstrates significant advantages in delineating the complete outlines and boundaries of large objects, such as the car and truck in the third row.

\subsection{Quantitative Results}

Table~\ref{tab:SemanticKITTI Quantitative Comparison} presents a comparison of our VLScene with other state-of-the-art camera-based SSC methods on the SemanticKITTI test sets. Our VLScene outperforms existing methods, achieving state-of-the-art results.
Compared to BRGScene, VLScene demonstrates an improvement of 2.16\% in mIoU and 1.8\% in IoU, with significant gains across all semantic categories. Additionally, compared to Symphonize, which extracts and fuses high-level instance features, VLScene achieves a lead of 2.95\% in IoU and 2.48\% in mIoU. These results validate the effectiveness of our approach in both geometric and semantic aspects, and VLScene achieves the highest mIoU in nearly all categories.

As shown in Table~\ref{tab:KITTI360Test Quantitative Comparison}, VLScene also exhibits a significant advantage in semantic and geometric analysis over current camera-based approaches on the rich data samples SSCBench-KITTI-360 benchmark, surpassing all published methods in both IoU and mIoU metrics.

Furthermore, Table~\ref{tab:diff range} shows that we provide different ranges of results on the SemanticKITTI validation set. It can be seen that our method significantly surpasses the existing methods at all three distances. 

In addition, as shown in Table~\ref{tab:cost}, we compare the inference time and number of parameters with other SOTA methods on the SemanticKITTI validation set. Our method achieves SOTA performance with 17.83\% mIoU using only 47.4M parameters. VLScene also demonstrates superior inference time while being more lightweight.


\begin{table}[t]
  \centering
  \small

  \begin{tabular}{l|l|ccc}
    \toprule
    \multirow{2}{*}{\textbf{Methods}}  & \multirow{2}{*}{\textbf{Venues}}& & \textbf{mIoU} &  \\
    & &12.8m &25.6m & 51.2m  \\
    \midrule
    SSCNet  &CVPR'17&16.32& 14.55& 10.27\\
    LMSCNet   &3DV'20&15.69& 14.13& 9.94\\
    MonoScene  & CVPR'22&12.25& 12.22& 11.30\\
    VoxFormer & CVPR'23 & 17.66& 16.48& 12.35\\
    OccFormer& ICCV'23 & 20.91& 17.90& 13.46\\
    HASSC& CVPR'24& 18.98 &17.95& 13.48\\
    H2GFormer& AAAI'24 & 20.49& 18.39& 13.73\\
    BRGScene& IJCAI'24 & 23.27& 21.15& 15.24\\
    \textbf{Ours} & &  \textbf{26.51}& \textbf{24.37}& \textbf{17.83}\\
    \bottomrule
  \end{tabular}
      \caption{Comparison of different ranges on SemanticKITTI val set.}
  \label{tab:diff range}
\end{table}
\begin{table}[t]
  \centering\small

  \begin{tabular}{l|ccc}
    \toprule
    {\textbf{Method}}& {\textbf{mIoU (\%)}$\uparrow$}&\textbf{Times (s)$\downarrow$}&\textbf{Params (M)$\downarrow$} \\
    \midrule
    MonoScene& 11.08&0.274&132.4\\
    OccFormer& {13.46}&0.338&203.4 \\
    VoxFormer& {13.35}&0.256&57.9 \\
    Symphonize& {14.89}&0.319&59.3 \\
    BRGScene& {15.43}&0.285&161.4 \\
    {Ours}&{\textbf{17.83}}&\textbf{0.233}&\textbf{47.4} \\
    \bottomrule
  \end{tabular}
    \caption{Comparison of inference time and number of parameters.}
  \label{tab:cost}
\end{table}

\begin{table}[t]
\centering\small
\setlength{\tabcolsep}{3pt}
  \begin{tabular}{c|cccc|ccc}
    \toprule
     \textbf{Method}&RepViT& {VLGD} & {NGP} & {SSI} & \textbf{IoU} & \textbf{mIoU}& \textbf{Params} \\

    \midrule
      Baseline&{}& {} & {} & {} &  {43.54} & {15.43}&161.4M \\
     1&{\checkmark}& {} & {} & {} &  {43.13} & {15.59}&120.0M \\
     2&{\checkmark}& {\checkmark} & {} & {} &  {44.02} & {16.58}&120.0M \\
     3&{\checkmark}& {\checkmark} & {\checkmark} & {} &  {44.58} & {16.43}&\textbf{35.6M} \\
     4&{\checkmark}& {\checkmark} & {} & {\checkmark} &  {44.03} & {17.21}&45.9M \\
     5&{\checkmark}& {} & {\checkmark} & {\checkmark} &  {44.15} & {16.66}&47.4M \\
     6&{\checkmark}&{\checkmark } & {\checkmark} & {\checkmark}& {\textbf{44.69}} &{\textbf{17.83}}&{47.4M} \\
    \bottomrule
\end{tabular}
  \caption{Ablation study for Architecture.}
  \label{tab:architecture}

\end{table}
\begin{table}[t]
  \centering\small
  \begin{tabular}{l|cc}
    \toprule
    {\textbf{Setting}} & {\textbf{IoU}} & {\textbf{mIoU}} \\
    \midrule
    Baseline  &  {43.54} & {15.43} \\
    Logits Distillation & {43.75} & {16.32} \\
    Feature Distillation & {43.88} & {16.54}  \\
    {VLGD (Ours)} & {\textbf{44.69}} & {\textbf{17.83}} \\
    \bottomrule
  \end{tabular}
  \caption{Ablation study for Vision-Language Guidance Distillation.}
  \label{tab:kd}
\end{table}

\begin{table}[!h]
  \centering\small
  \begin{tabular}{l|ccc}
    \toprule
    {\textbf{Setting}} & {\textbf{IoU}} & {\textbf{mIoU}}&\textbf{Params} \\
    \midrule
    3D Swin & 44.18 & 15.87 &82.3M\\
    3D ResNet & 44.03 & 15.66 & 37.4M \\
    {GSSA (Ours)} & {\textbf{44.69}} & {\textbf{17.83}} & \textbf{13.3M}\\
    \bottomrule
  \end{tabular}
  \caption{Ablation study for Geometric-Semantic Sparse Aware.}
  \label{tab:gssa}
\end{table}

\subsection{Ablation Studies}
\paragraph{Ablation on the Architectural Components.}
The Table~\ref{tab:architecture} shows a breakdown analysis of various architectural components in the VLScene, including the lightweight image encoder RepViT, vision-language guidance distillation (VLGD), neighborhood geometric propagation (NGP), and sparse semantic interaction (SSI). We will analyze the contents of the Table~\ref{tab:architecture} line by line. First, we adopted BRGScene~\cite{li2023stereoscene} as our baseline. (1) When we replaced the baseline image encoder with RepViT, the model parameters were reduced by 41.4M. (2) Significantly improved mIoU by 0.99\% when equipped with our VLGD, proving that VLGD injects rich semantic information through knowledge distillation. (3) The introduction of NGP can effectively improve its geometric completion performance, and the IoU has achieved a 0.56\% improvement, while only 35.6M parameters are required. (4) The SSI module enhanced semantic information through contextual interaction, and the mIoU increased by 0.63\%. (5) The combination of NGP and SSI improves both geometric and semantic performance. (6) The final full model uses only 47.4M parameters, achieving 1.15\% IoU and 2.4\% mIoU improvements over baseline, proving that all of these components contribute to the best results.
\paragraph{Ablation Study for VLGD.}
To further explore the role of VLGD, we examine the individual effects of feature distillation and logical distillation, as shown in Table~\ref{tab:kd}. Our findings indicate that visual language knowledge distillation significantly enhances the model's semantic representation learning. Both distillation methods result in improvements in IoU and mIoU, with VLGD contributing to a 1.15\% increase in IoU and a 2.4\% rise in mIoU compared to the baseline~\cite{li2023stereoscene}. Notably, our method introduces no additional time or computational overhead during inference.

\paragraph{Ablation Study for GSSA.}
As shown in Table~\ref{tab:gssa}, We compare GSSA with other baseline methods. The fixed size of the convolutional kernel in 3D ResNet~\cite{he2016resnet} limits its ability to capture information over a broader region. Similarly, the restricted range of local windows in 3D Swin~\cite{liu2022swin3d} challenges its capacity to effectively capture telematics through self-attention, with sliding window operations only impacting a small neighborhood around each window. Moreover, these methods require more parameters and computational resources. Our GSSA achieves significant improvements while utilizing only 13.3M parameters.

\section{Conclusion}
\label{sec:conclusion}
In this paper, we propose a novel method, VLScene: Vision-Language Guidance Distillation for Camera-based 3D Semantic Scene Completion. 
Specifically, we design a vision-language guidance distillation process that effectively captures semantic knowledge from the surrounding environment and improves spatial context reasoning. In addition, we introduce a geometric-semantic sparse awareness mechanism to propagate geometric structures in the neighborhood and enhance semantic information through contextual sparse interactions.
Experimental results demonstrate that VLScene achieves SOTA performance on the SemanticKITTI and SSCBench-KITTI-360 datasets.
\appendix
\section*{Appendix Overview}
This technical appendix consists of the following sections:
\begin{itemize}
\item
We provide a setup of the VLScene method.
\item
We provide quantitative results from more experiments.
\item
We present more visual qualitative results of the SemanticKITTI val set.
\item
We analyze the shortcomings of our method and directions for future work.
\end{itemize} 

\section{Experimental Setup}
\label{detail}
\subsection{Datasets} 
\paragraph{SemanticKITTI.} The SemanticKITTI~\cite{behley2019semantickitti,Geiger2012kitti} dataset has dense semantic scene completion annotations and labels a voxelized scene with 20 semantic classes. The dataset comprises 10 training sequences, 1 validation sequence, and 11 testing sequences. The RGB images are resized to $1280 \times 384$ dimensions to serve as inputs. 

\paragraph{SSCBench-KITTI-360.} The SSCBench-KITTI-360~\cite{li2023sscbench,Liao2022kitti360} dataset provides 7 training sequences, 1 validation sequence, and 1 testing sequence, covering a total of 19 semantic classes. The RGB images are resized to $1408 \times 384$ resolution for input processing.

\begin{table*}[!ht]
  \centering
  \small
  \setlength{\tabcolsep}{2pt}
  \begin{tabular}{l|c|ccccccccccccccccccc|c}
    \toprule
    \textbf{Methods} &\textbf{IoU}   
& \rotatebox{90}{\vcenteredbox{\colorbox[RGB]{255,0,255}{\textcolor[RGB]{255,0,255}{\rule{1px}{1px}}}} \textbf{road} (15.30$\%$)} 
    & \rotatebox{90}{\vcenteredbox{\colorbox[RGB]{75,0,75}{\textcolor[RGB]{75,0,75}{\rule{1px}{1px}}}} \textbf{sidewalk} (11.13$\%$)} 
    & \rotatebox{90}{\vcenteredbox{\colorbox[RGB]{255,150,255}{\textcolor[RGB]{255,150,255}{\rule{1px}{1px}}}} \textbf{parking} (1.12$\%$)} 
    & \rotatebox{90}{\vcenteredbox{\colorbox[RGB]{175,0,75}{\textcolor[RGB]{175,0,75}{\rule{1px}{1px}}}} \textbf{other-grnd} (0.56$\%$)} 
    &  \rotatebox{90}{\vcenteredbox{\colorbox[RGB]{255,200,0}{\textcolor[RGB]{255,200,0}{\rule{1px}{1px}}}} \textbf{building} (14.10$\%$)} 
    &  \rotatebox{90}{\vcenteredbox{\colorbox[RGB]{100,150,245}{\textcolor[RGB]{100,150,245}{\rule{1px}{1px}}}} \textbf{car} (3.92$\%$)} 
    & \rotatebox{90}{\vcenteredbox{\colorbox[RGB]{80,30,180}{\textcolor[RGB]{80,30,180}{\rule{1px}{1px}}}} \textbf{truck} (0.16$\%$)} 
    & \rotatebox{90}{\vcenteredbox{\colorbox[RGB]{100,230,245}{\textcolor[RGB]{100,230,245}{\rule{1px}{1px}}}} \textbf{bicycle} (0.03$\%$)}  
    & \rotatebox{90}{\vcenteredbox{\colorbox[RGB]{30,60,150}{\textcolor[RGB]{30,60,150}{\rule{1px}{1px}}}} \textbf{motocycle} (0.03$\%$)} 
    & \rotatebox{90}{\vcenteredbox{\colorbox[RGB]{0,0,255}{\textcolor[RGB]{0,0,255}{\rule{1px}{1px}}}} \textbf{other-vehicle} (0.20$\%$)} 
    & \rotatebox{90}{\vcenteredbox{\colorbox[RGB]{0,175,0}{\textcolor[RGB]{0,175,0}{\rule{1px}{1px}}}} \textbf{vegetation} (39.3$\%$)} 
    & \rotatebox{90}{\vcenteredbox{\colorbox[RGB]{135,60,0}{\textcolor[RGB]{135,60,0}{\rule{1px}{1px}}}}  \textbf{trunk} (0.51$\%$)} 
    & \rotatebox{90}{\vcenteredbox{\colorbox[RGB]{150,240,80}{\textcolor[RGB]{150,240,80}{\rule{1px}{1px}}}} \textbf{terrain} (9.17$\%$)} 
    & \rotatebox{90}{\vcenteredbox{\colorbox[RGB]{255,30,30}{\textcolor[RGB]{255,30,30}{\rule{1px}{1px}}}} \textbf{person} (0.07$\%$)} 
    & \rotatebox{90}{\vcenteredbox{\colorbox[RGB]{255,40,200}{\textcolor[RGB]{255,40,200}{\rule{1px}{1px}}}} \textbf{bicylist} (0.07$\%$)} 
    & \rotatebox{90}{\vcenteredbox{\colorbox[RGB]{150,30,90}{\textcolor[RGB]{150,30,90}{\rule{1px}{1px}}}}  \textbf{motorcyclist} (0.05$\%$)} 
    &  \rotatebox{90}{\vcenteredbox{\colorbox[RGB]{255,120,50}{\textcolor[RGB]{255,120,50}{\rule{1px}{1px}}}} \textbf{fence} (3.90$\%$)} 
    & \rotatebox{90}{\vcenteredbox{\colorbox[RGB]{255,240,150}{\textcolor[RGB]{255,240,150}{\rule{1px}{1px}}}} \textbf{pole} (0.29$\%$)} 
    & \rotatebox{90}{\vcenteredbox{\colorbox[RGB]{255,0,0}{\textcolor[RGB]{255,0,0}{\rule{1px}{1px}}}} \textbf{traf.-sign} (0.08$\%$)}
    & \textbf{mIoU}  \\
    
    \midrule
    MonoScene& 36.86&56.5&26.7&14.3&0.5&14.1&23.3&7.0&0.6&0.5&1.5&17.9&2.8&29.6&1.9&1.2&0.0&5.8&4.1&2.3&11.08 \\
    
    TPVFormer& 35.61&56.5&25.9&20.6&0.9&13.9&23.8&8.1&0.4&0.0&4.4&16.9&2.3&30.4&0.5&0.9&0.0&5.9&3.1&1.5&11.36 \\
    
    OccFormer& 36.50  &58.9&26.9&19.6&0.3&14.4&25.1&25.5&0.8&1.2&8.5&19.6&3.9&32.6&2.8&2.8&0.0&5.6&4.3&2.9& 13.46\\

    VoxFormer& 44.15 &53.6&26.5&19.7&0.4&19.5&26.5&7.3&1.3&0.6&7.8&26.1&6.1&33.0&1.9&2.0&0.0&7.3&9.2&4.9& 13.35\\
    H2GFormer & 44.57&56.1&29.1&17.8&0.5&19.7&28.2&10.0&0.5&0.5&7.4&26.3&6.8&34.4&1.5&2.9&0.0&7.2&7.9&4.7&13.73\\
    HASSC&\textbf{44.82}&57.1&28.3&15.9&\textbf{1.0}&19.1&27.2&9.9&0.9&0.9&5.6&25.5&6.2&32.9&2.8&\textbf{4.7}&0.0&6.6&7.7&4.1&13.48\\
    Symphonize&41.92&56.4&27.6&15.3&0.9&21.6&28.7&20.4&2.5&2.8&13.9&25.7&6.6&30.9&3.5&2.2&0.0&8.4&9.6&5.8&14.89\\
    OctOcc&44.02&55.1&27.9&22.6&0.5&20.3&27.8&6.0&\textbf{2.6}&2.0&6.8&\textbf{26.6}&6.8&33.8&2.7&0.0&0.0&8.9&9.3&5.6&14.59\\


    \rowcolor{gray!20}\textbf{Ours}  & {44.69} &{\textbf{63.1}}&{\textbf{31.1}}& \textbf{24.4}&0.2& {\textbf{24.9}}& {\textbf{33.4}}&\textbf{30.7}&1.8& {\textbf{3.6}}&\textbf{18.3}& 26.0& {\textbf{8.1}}& {\textbf{35.3}}&\textbf{4.3}& 2.6& {\textbf{0.0}}& {\textbf{12.1}}& {\textbf{11.9}}& {\textbf{6.3}}&  {\textbf{17.83}}\\
    \bottomrule
  \end{tabular}
    \caption{Quantitative results on the SemanticKITTI validation set. {\textbf{Bold}} denotes the best performance.}
  \label{tab:SemanticKITTI Quantitative Comparison}
\end{table*}

\subsection{Metrics} Following \cite{song2017semantic}, the main consideration of SSC is the mean Intersection over Union (mIoU), which considers the IoU of all semantic classes for prediction without considering the free space. The mIoU is calculated by:
\begin{equation}
  mIoU = \frac{1}{C} \sum^{C}_{c=1}\frac{TP_c}{TN_c+FP_c+FN_c}
\end{equation}
Here, $TP_c$, $TN_c$, $FP_c$, and $FN_c$ are the true positives, true negatives, false positives and false negatives predictions for class $c$.

The SSC task requires considering pure geometric reconstruction quality, while mIoU takes into account all semantic categories. Therefore, Intersection over Union (IoU), Precision, and Recall are often used to represent scene representations indicating empty or occupied areas.
\subsection{Implementation Details} 

\paragraph{Network architecture.} In contrast to previous studies~\cite{cao2022monoscene,li2023stereoscene,zhang2023occformer}, we use a more lightweight neural network RepViT~\cite{wang2023repvit} as the image backbone. We also use the visual language model LSeg~\cite{radford2021clip,li2022lseg} to extract visual and language features. For the view transformation, we use LSS for 2D-3D projection to generate a 3D feature volume of $128 \times 128 \times 16$ and 128 channels. The NGP module contains a $7\times7\times7$ 3D convolution layer and a $3\times3\times3$ 3D convolution layer. The SSI module contains two downsampling layers and two upsampling layers consisting of asymmetric residual blocks.
\paragraph{Training settings.} The final output of SemantiKITTI is 20 categories, and that of SSBench-KITTI-360 is 19 categories. The scene size of all datasets is $51.2m \times 51.2m \times 64m$, and the voxel grid size is $256 \times 256 \times 32$. By default, the models are trained for 30 epochs. We use the AdamW~\cite{loshchilov2017decoupled} optimizer to optimize the process with an initial learning rate of 1e-4 and a weight decay of 0.01. We also adopt a multi-step scheduler to reduce the learning rate. All models are trained on two A100 Nvidia GPUs with 80G memory and batch size 4.

\section{Quantitative Results}
\label{quantitative}
\paragraph{Quantitative Comparison in SemanticKITTI validation set.}
To provide a more thorough comparison, we provide additional quantitative results of semantic scene completion (SSC) on the SemanticKITTI validation set in Tab \ref{tab:SemanticKITTI Quantitative Comparison}. The results further demonstrate the effectiveness of our approach in enhancing 3D scene perception performance.

Compared with the previous state-of-the-art methods, VLScene is far superior to other OctOcc~\cite{ouyang2024octocc} in semantic scene understanding, with a 3.24\% increase in mIoU. In addition, compared with Symphonize~\cite{jiang2024symphonize}, huge improvements are made in both occupancy and semantics. IoU and mIoU enhancement are of great significance for practical applications. It proves that we are not simply reducing a certain metric to achieve semantic scene completion.

\section{Visualization Results}
\label{Visualizations}
\begin{figure*}[ht]
\centering
  \includegraphics[width=\textwidth]{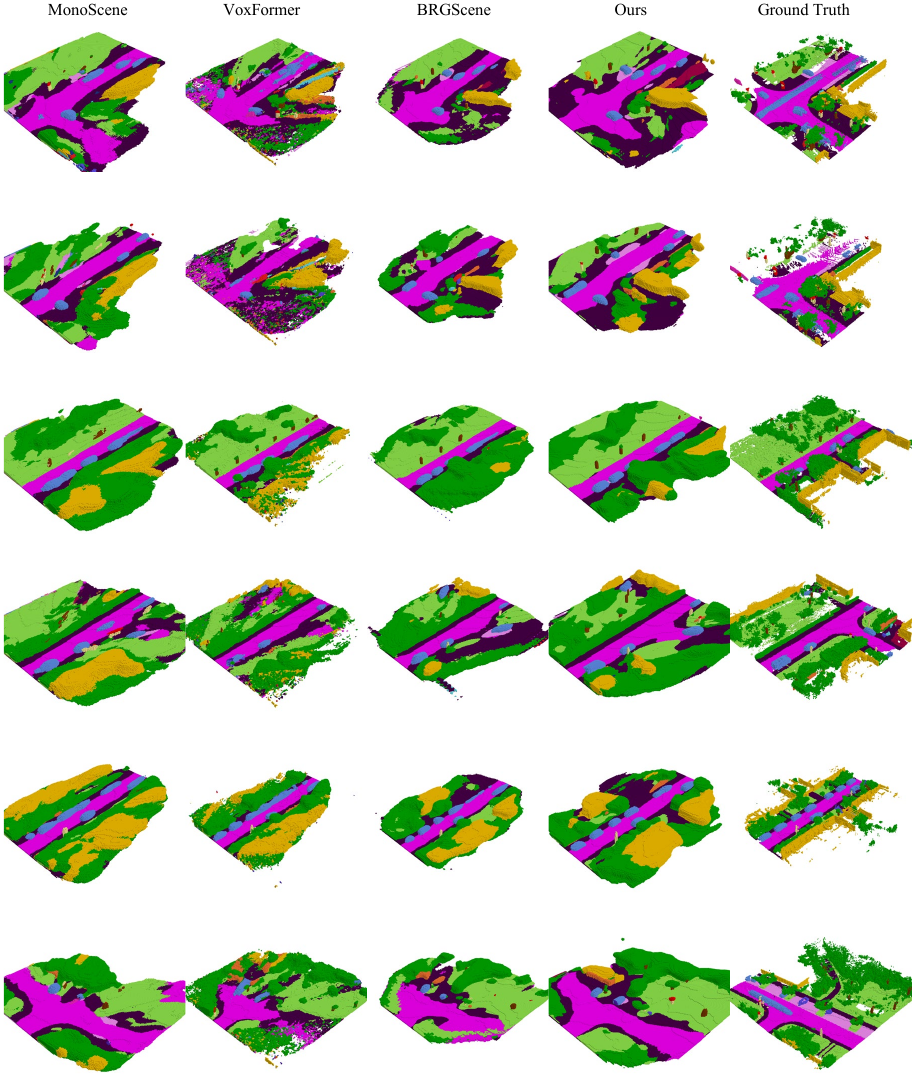}
  \caption{Qualitative results on the SemanticKITTI validation set.}
  \label{fig:sup1}
\end{figure*}

\begin{figure*}[ht]
\centering
  \includegraphics[width=\textwidth]{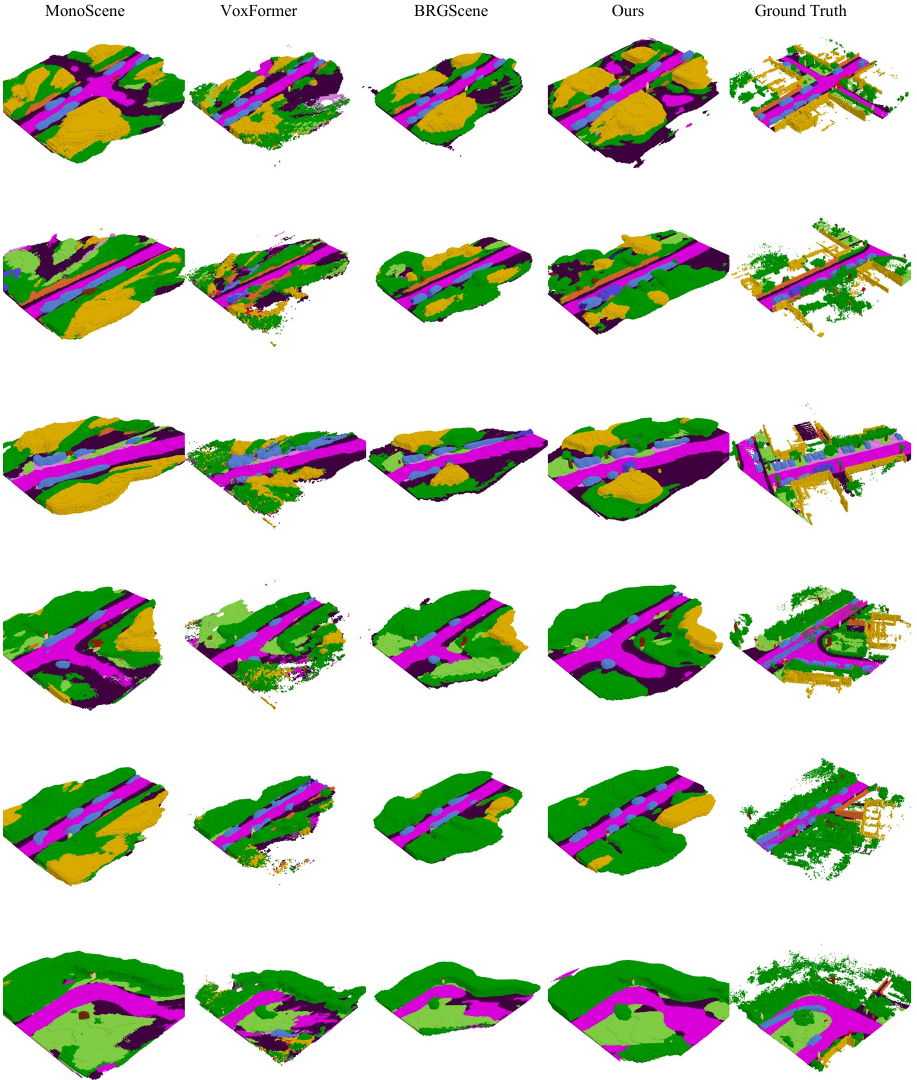}
  \caption{Qualitative results on the SemanticKITTI validation set.}
  \label{fig:sup2}
\end{figure*}

We report the performance of more visual results on the SemanticKITTI validation set in Figures ~\ref{fig:sup1} and ~\ref{fig:sup2}. We compare with MonoScene~\cite{cao2022monoscene}, VoxFormer~\cite{li2023voxformer}, and BRGScene~\cite{li2023stereoscene}. In general, our method performs more fine-grained segmentation of the scene and maintains clear segmentation boundaries. For example, in the segmentation completion results of cars, we predict clear separation for each car. In contrast, other methods show continuous semantic errors for occluded cars. In addition, our VLScene can effectively capture the location and details of small objects (such as persons, poles, traffic signs, etc.). Finally, surprisingly, our method can also show excellent scene hallucination performance in areas outside the camera's field of view. We also provide a video in the appendix to more intuitively demonstrate the performance.
\section{Discussions}
\label{sec:limit}
VLScene shows strong performance on benchmarks and the number of parameters and inference time of the model are further improved. This is beneficial for the deployment of real-world autonomous driving applications. Semantic scene completion in multi-camera settings is also worth attention, which we leave for our future work. Meanwhile, the legal challenges of autonomous driving and the privacy and data security risks are still topics of debate.

\section{Acknowledgments}
The work is supported by the Science and Technology Innovation 2030 - New Generation Artificial Intelligence Major Project (2021ZD40300), the National Natural Science Foundation of China (Grant Nos.62225205, 62202151, 52203288, 62473137, 62321003), the China Association for Science and Technology Young Talent Support Project Doctoral Special Program, and the Postgraduate Scientific Research Innovation Project of Hunan Province (CX20240427).

\bibliography{aaai25}

\begin{thebibliography}{48}
\providecommand{\natexlab}[1]{#1}

\bibitem[{Behley et~al.(2019)Behley, Garbade, Milioto, Quenzel, Behnke, Stachniss, and Gall}]{behley2019semantickitti}
Behley, J.; Garbade, M.; Milioto, A.; Quenzel, J.; Behnke, S.; Stachniss, C.; and Gall, J. 2019.
\newblock Semantickitti: A dataset for semantic scene understanding of lidar sequences.
\newblock In \emph{ICCV}, 9297--9307.

\bibitem[{Cai et~al.(2021)Cai, Chen, Zhang, Lin, Wang, and Li}]{cai2021semantic}
Cai, Y.; Chen, X.; Zhang, C.; Lin, K.-Y.; Wang, X.; and Li, H. 2021.
\newblock Semantic scene completion via integrating instances and scene in-the-loop.
\newblock In \emph{CVPR}, 324--333.

\bibitem[{Cao and de~Charette(2022)}]{cao2022monoscene}
Cao, A.-Q.; and de~Charette, R. 2022.
\newblock MonoScene: Monocular 3D Semantic Scene Completion.
\newblock In \emph{CVPR}.

\bibitem[{Chen et~al.(2023)Chen, Liu, Kong, Zhu, Ma, Li, Hou, Qiao, and Wang}]{chen2023clip2scene}
Chen, R.; Liu, Y.; Kong, L.; Zhu, X.; Ma, Y.; Li, Y.; Hou, Y.; Qiao, Y.; and Wang, W. 2023.
\newblock Clip2scene: Towards label-efficient 3d scene understanding by clip.
\newblock In \emph{Proceedings of the IEEE/CVF Conference on Computer Vision and Pattern Recognition}, 7020--7030.

\bibitem[{Chen et~al.(2022)Chen, Li, Zhang, Fang, Jiang, and Zhao}]{chen2022bevdistill}
Chen, Z.; Li, Z.; Zhang, S.; Fang, L.; Jiang, Q.; and Zhao, F. 2022.
\newblock Bevdistill: Cross-modal bev distillation for multi-view 3d object detection.
\newblock \emph{arXiv preprint arXiv:2211.09386}.

\bibitem[{Dong et~al.(2023)Dong, Bao, Zheng, Zhang, Chen, Yang, Zeng, Zhang, Yuan, Chen et~al.}]{dong2023maskclip}
Dong, X.; Bao, J.; Zheng, Y.; Zhang, T.; Chen, D.; Yang, H.; Zeng, M.; Zhang, W.; Yuan, L.; Chen, D.; et~al. 2023.
\newblock Maskclip: Masked self-distillation advances contrastive language-image pretraining.
\newblock In \emph{Proceedings of the IEEE/CVF Conference on Computer Vision and Pattern Recognition}, 10995--11005.

\bibitem[{Geiger, Lenz, and Urtasun(2012)}]{Geiger2012kitti}
Geiger, A.; Lenz, P.; and Urtasun, R. 2012.
\newblock Are we ready for Autonomous Driving? The KITTI Vision Benchmark Suite.
\newblock In \emph{CVPR}.

\bibitem[{Ghiasi et~al.(2022)Ghiasi, Gu, Cui, and Lin}]{ghiasi2022openseg}
Ghiasi, G.; Gu, X.; Cui, Y.; and Lin, T.-Y. 2022.
\newblock Scaling open-vocabulary image segmentation with image-level labels.
\newblock In \emph{European Conference on Computer Vision}, 540--557. Springer.

\bibitem[{Guo and Tong(2018)}]{guo2018view}
Guo, Y.; and Tong, X. 2018.
\newblock View-volume network for semantic scene completion from a single depth image.
\newblock In \emph{IJCAI}, 726--732.

\bibitem[{He et~al.(2016)He, Zhang, Ren, and Sun}]{he2016resnet}
He, K.; Zhang, X.; Ren, S.; and Sun, J. 2016.
\newblock Deep residual learning for image recognition.
\newblock In \emph{CVPR}, 770--778.

\bibitem[{Hinton, Vinyals, and Dean(2015)}]{hinton2015distilling}
Hinton, G.; Vinyals, O.; and Dean, J. 2015.
\newblock Distilling the knowledge in a neural network.
\newblock \emph{arXiv preprint arXiv:1503.02531}.

\bibitem[{Hong, Dai, and Ding(2022)}]{hong2022cmkd}
Hong, Y.; Dai, H.; and Ding, Y. 2022.
\newblock Cross-modality knowledge distillation network for monocular 3d object detection.
\newblock In \emph{European Conference on Computer Vision}, 87--104. Springer.

\bibitem[{Huang et~al.(2023)Huang, Zheng, Zhang, Zhou, and Lu}]{huang2023tri}
Huang, Y.; Zheng, W.; Zhang, Y.; Zhou, J.; and Lu, J. 2023.
\newblock Tri-perspective view for vision-based 3d semantic occupancy prediction.
\newblock In \emph{CVPR}, 9223--9232.

\bibitem[{Jiang et~al.(2024)Jiang, Cheng, Gao, Zhang, Lin, Liu, and Wang}]{jiang2024symphonize}
Jiang, H.; Cheng, T.; Gao, N.; Zhang, H.; Lin, T.; Liu, W.; and Wang, X. 2024.
\newblock Symphonize 3d semantic scene completion with contextual instance queries.
\newblock In \emph{CVPR}, 20258--20267.

\bibitem[{Kuo et~al.(2023)Kuo, Cui, Gu, Piergiovanni, and Angelova}]{kuo2023F-VLM}
Kuo, W.; Cui, Y.; Gu, X.; Piergiovanni, A.; and Angelova, A. 2023.
\newblock Open-Vocabulary Object Detection upon Frozen Vision and Language Models.
\newblock In \emph{The Eleventh International Conference on Learning Representations}.

\bibitem[{Li et~al.(2023{\natexlab{a}})Li, Sun, Jin, Zeng, Zhu, Wang, Zhang, Okae, Xiao, and Du}]{li2023stereoscene}
Li, B.; Sun, Y.; Jin, X.; Zeng, W.; Zhu, Z.; Wang, X.; Zhang, Y.; Okae, J.; Xiao, H.; and Du, D. 2023{\natexlab{a}}.
\newblock Stereoscene: Bev-assisted stereo matching empowers 3d semantic scene completion.
\newblock \emph{arXiv preprint arXiv:2303.13959}.

\bibitem[{Li et~al.(2022)Li, Weinberger, Belongie, Koltun, and Ranftl}]{li2022lseg}
Li, B.; Weinberger, K.~Q.; Belongie, S.; Koltun, V.; and Ranftl, R. 2022.
\newblock Language-driven semantic segmentation.
\newblock \emph{arXiv preprint arXiv:2201.03546}.

\bibitem[{Li et~al.(2020)Li, Han, Wang, Liu, and Yuan}]{li2020aicnet}
Li, J.; Han, K.; Wang, P.; Liu, Y.; and Yuan, X. 2020.
\newblock Anisotropic convolutional networks for 3d semantic scene completion.
\newblock In \emph{CVPR}, 3351--3359.

\bibitem[{Li et~al.(2019)Li, Liu, Gong, Shi, Yuan, Zhao, and Reid}]{li2019rgbd}
Li, J.; Liu, Y.; Gong, D.; Shi, Q.; Yuan, X.; Zhao, C.; and Reid, I. 2019.
\newblock Rgbd based dimensional decomposition residual network for 3d semantic scene completion.
\newblock In \emph{CVPR}, 7693--7702.

\bibitem[{Li et~al.(2023{\natexlab{b}})Li, Li, Liu, Gong, Li, Chen, Wang, Li, Jiang, Yu, Wang, Zhao, Yu, and Feng}]{li2023sscbench}
Li, Y.; Li, S.; Liu, X.; Gong, M.; Li, K.; Chen, N.; Wang, Z.; Li, Z.; Jiang, T.; Yu, F.; Wang, Y.; Zhao, H.; Yu, Z.; and Feng, C. 2023{\natexlab{b}}.
\newblock SSCBench: Monocular 3D Semantic Scene Completion Benchmark in Street Views.
\newblock \emph{arXiv preprint arXiv:2306.09001}.

\bibitem[{Li et~al.(2023{\natexlab{c}})Li, Yu, Choy, Xiao, Alvarez, Fidler, Feng, and Anandkumar}]{li2023voxformer}
Li, Y.; Yu, Z.; Choy, C.; Xiao, C.; Alvarez, J.~M.; Fidler, S.; Feng, C.; and Anandkumar, A. 2023{\natexlab{c}}.
\newblock VoxFormer: Sparse Voxel Transformer for Camera-based 3D Semantic Scene Completion.
\newblock In \emph{CVPR}.

\bibitem[{Liao, Xie, and Geiger(2022)}]{Liao2022kitti360}
Liao, Y.; Xie, J.; and Geiger, A. 2022.
\newblock {KITTI}-360: A Novel Dataset and Benchmarks for Urban Scene Understanding in 2D and 3D.
\newblock \emph{TPAMI}.

\bibitem[{Lin et~al.(2017)Lin, Doll{\'a}r, Girshick, He, Hariharan, and Belongie}]{lin2017feature}
Lin, T.-Y.; Doll{\'a}r, P.; Girshick, R.; He, K.; Hariharan, B.; and Belongie, S. 2017.
\newblock Feature pyramid networks for object detection.
\newblock In \emph{CVPR}, 2117--2125.

\bibitem[{Liu et~al.(2021)Liu, Huang, Chiang, Su, Liu, Chen, Tseng, and Hsu}]{liu2021PPKT}
Liu, Y.-C.; Huang, Y.-K.; Chiang, H.-Y.; Su, H.-T.; Liu, Z.-Y.; Chen, C.-T.; Tseng, C.-Y.; and Hsu, W.~H. 2021.
\newblock Learning from 2d: Contrastive pixel-to-point knowledge transfer for 3d pretraining.
\newblock \emph{arXiv preprint arXiv:2104.04687}.

\bibitem[{Liu et~al.(2022)Liu, Ning, Cao, Wei, Zhang, Lin, and Hu}]{liu2022swin3d}
Liu, Z.; Ning, J.; Cao, Y.; Wei, Y.; Zhang, Z.; Lin, S.; and Hu, H. 2022.
\newblock Video swin transformer.
\newblock In \emph{CVPR}, 3202--3211.

\bibitem[{Loshchilov and Hutter(2017)}]{loshchilov2017decoupled}
Loshchilov, I.; and Hutter, F. 2017.
\newblock Decoupled weight decay regularization.
\newblock \emph{arXiv preprint arXiv:1711.05101}.

\bibitem[{Mirzadeh et~al.(2020)Mirzadeh, Farajtabar, Li, Levine, Matsukawa, and Ghasemzadeh}]{mirzadeh2020improved}
Mirzadeh, S.~I.; Farajtabar, M.; Li, A.; Levine, N.; Matsukawa, A.; and Ghasemzadeh, H. 2020.
\newblock Improved knowledge distillation via teacher assistant.
\newblock In \emph{Proceedings of the AAAI conference on artificial intelligence}, volume~34, 5191--5198.

\bibitem[{Ouyang et~al.(2024)Ouyang, Song, Feng, and Xu}]{ouyang2024octocc}
Ouyang, W.; Song, X.; Feng, B.; and Xu, Z. 2024.
\newblock OctOcc: High-Resolution 3D Occupancy Prediction with Octree.
\newblock In \emph{Proceedings of the AAAI Conference on Artificial Intelligence}, volume~38, 4369--4377.

\bibitem[{Peng et~al.(2023)Peng, Genova, Jiang, Tagliasacchi, Pollefeys, and Funkhouser}]{Peng2023OpenScene}
Peng, S.; Genova, K.; Jiang, C.~M.; Tagliasacchi, A.; Pollefeys, M.; and Funkhouser, T. 2023.
\newblock OpenScene: 3D Scene Understanding with Open Vocabularies.

\bibitem[{Radford et~al.(2021)Radford, Kim, Hallacy, Ramesh, Goh, Agarwal, Sastry, Askell, Mishkin, Clark et~al.}]{radford2021clip}
Radford, A.; Kim, J.~W.; Hallacy, C.; Ramesh, A.; Goh, G.; Agarwal, S.; Sastry, G.; Askell, A.; Mishkin, P.; Clark, J.; et~al. 2021.
\newblock Learning transferable visual models from natural language supervision.
\newblock In \emph{International conference on machine learning}, 8748--8763. PMLR.

\bibitem[{Rist et~al.(2021)Rist, Emmerichs, Enzweiler, and Gavrila}]{rist2021semantic}
Rist, C.~B.; Emmerichs, D.; Enzweiler, M.; and Gavrila, D.~M. 2021.
\newblock Semantic scene completion using local deep implicit functions on lidar data.
\newblock \emph{TPAMI}, 44(10): 7205--7218.

\bibitem[{Roldao, de~Charette, and Verroust-Blondet(2020)}]{roldao2020lmscnet}
Roldao, L.; de~Charette, R.; and Verroust-Blondet, A. 2020.
\newblock Lmscnet: Lightweight multiscale 3d semantic completion.
\newblock In \emph{3DV}, 111--119. IEEE.

\bibitem[{Song et~al.(2017)Song, Yu, Zeng, Chang, Savva, and Funkhouser}]{song2017semantic}
Song, S.; Yu, F.; Zeng, A.; Chang, A.~X.; Savva, M.; and Funkhouser, T. 2017.
\newblock Semantic scene completion from a single depth image.
\newblock In \emph{CVPR}, 1746--1754.

\bibitem[{Wang et~al.(2023)Wang, Chen, Lin, Han, and Ding}]{wang2023repvit}
Wang, A.; Chen, H.; Lin, Z.; Han, J.; and Ding, G. 2023.
\newblock RepViT: Revisiting Mobile CNN From ViT Perspective.
\newblock arXiv:2307.09283.

\bibitem[{Wang et~al.(2022)Wang, Li, Liao, Jiang, Wu, Wang, Qian, and Liu}]{wang2022head}
Wang, L.; Li, X.; Liao, Y.; Jiang, Z.; Wu, J.; Wang, F.; Qian, C.; and Liu, S. 2022.
\newblock Head: Hetero-assists distillation for heterogeneous object detectors.
\newblock In \emph{European Conference on Computer Vision}, 314--331. Springer.

\bibitem[{Wang et~al.(2024)Wang, Yu, Li, Liu, Liu, Chen, and Zhu}]{wang2024HASSC}
Wang, S.; Yu, J.; Li, W.; Liu, W.; Liu, X.; Chen, J.; and Zhu, J. 2024.
\newblock Not all voxels are equal: Hardness-aware semantic scene completion with self-distillation.
\newblock In \emph{CVPR}, 14792--14801.

\bibitem[{Wang and Tong(2024)}]{wang2024h2gformer}
Wang, Y.; and Tong, C. 2024.
\newblock H2gformer: Horizontal-to-global voxel transformer for 3d semantic scene completion.
\newblock In \emph{Proceedings of the AAAI Conference on Artificial Intelligence}, volume~38, 5722--5730.

\bibitem[{Xia et~al.(2023)Xia, Liu, Li, Zhu, Ma, Li, Hou, and Qiao}]{Xia_2023_scpnet}
Xia, Z.; Liu, Y.; Li, X.; Zhu, X.; Ma, Y.; Li, Y.; Hou, Y.; and Qiao, Y. 2023.
\newblock SCPNet: Semantic Scene Completion on Point Cloud.
\newblock In \emph{CVPR}, 17642--17651.

\bibitem[{Xue et~al.(2024)Xue, Li, Wu, Tang, Li, and Duan}]{xue2024bi}
Xue, Y.; Li, R.; Wu, F.; Tang, Z.; Li, K.; and Duan, M. 2024.
\newblock Bi-SSC: Geometric-Semantic Bidirectional Fusion for Camera-based 3D Semantic Scene Completion.
\newblock In \emph{Proceedings of the IEEE/CVF Conference on Computer Vision and Pattern Recognition}, 20124--20134.

\bibitem[{Yan et~al.(2021)Yan, Gao, Li, Zhang, Li, Huang, and Cui}]{yan2021sparse}
Yan, X.; Gao, J.; Li, J.; Zhang, R.; Li, Z.; Huang, R.; and Cui, S. 2021.
\newblock Sparse single sweep lidar point cloud segmentation via learning contextual shape priors from scene completion.
\newblock In \emph{AAAI}, volume~35, 3101--3109.

\bibitem[{Yan et~al.(2022)Yan, Gao, Zheng, Zheng, Zhang, Cui, and Li}]{yan20222dpass}
Yan, X.; Gao, J.; Zheng, C.; Zheng, C.; Zhang, R.; Cui, S.; and Li, Z. 2022.
\newblock 2dpass: 2d priors assisted semantic segmentation on lidar point clouds.
\newblock In \emph{European Conference on Computer Vision}, 677--695. Springer.

\bibitem[{Yang et~al.(2023)Yang, Ding, Deng, Wang, and Qi}]{yang2023regionplc}
Yang, J.; Ding, R.; Deng, W.; Wang, Z.; and Qi, X. 2023.
\newblock Regionplc: Regional point-language contrastive learning for open-world 3d scene understanding.
\newblock \emph{arXiv preprint arXiv:2304.00962}.

\bibitem[{Yao et~al.(2023)Yao, Li, Sun, Cai, Li, Ouyang, and Li}]{yao2023ndc}
Yao, J.; Li, C.; Sun, K.; Cai, Y.; Li, H.; Ouyang, W.; and Li, H. 2023.
\newblock Ndc-scene: Boost monocular 3d semantic scene completion in normalized device coordinates space.
\newblock In \emph{ICCV}, 9455--9465.

\bibitem[{Zhang et~al.(2018)Zhang, Zhao, Yao, Chen, Zhang, and Liao}]{zhang2018efficient}
Zhang, J.; Zhao, H.; Yao, A.; Chen, Y.; Zhang, L.; and Liao, H. 2018.
\newblock Efficient semantic scene completion network with spatial group convolution.
\newblock In \emph{ECCV}, 733--749.

\bibitem[{Zhang, Zhu, and Du(2023)}]{zhang2023occformer}
Zhang, Y.; Zhu, Z.; and Du, D. 2023.
\newblock OccFormer: Dual-path Transformer for Vision-based 3D Semantic Occupancy Prediction.
\newblock \emph{arXiv preprint arXiv:2304.05316}.

\bibitem[{Zheng et~al.(2024)Zheng, Li, Li, Zheng, Jin, Zhong, Long, Zhao, and Zhang}]{zheng2024monoocc}
Zheng, Y.; Li, X.; Li, P.; Zheng, Y.; Jin, B.; Zhong, C.; Long, X.; Zhao, H.; and Zhang, Q. 2024.
\newblock Monoocc: Digging into monocular semantic occupancy prediction.
\newblock \emph{arXiv preprint arXiv:2403.08766}.

\bibitem[{Zhou et~al.(2023)Zhou, Liu, Hu, Zhou, and Ma}]{zhou2023unidistill}
Zhou, S.; Liu, W.; Hu, C.; Zhou, S.; and Ma, C. 2023.
\newblock UniDistill: A Universal Cross-Modality Knowledge Distillation Framework for 3D Object Detection in Bird's-Eye View.
\newblock In \emph{Proceedings of the IEEE/CVF Conference on Computer Vision and Pattern Recognition}, 5116--5125.

\bibitem[{Zhu et~al.(2021)Zhu, Zhou, Wang, Hong, Ma, Li, Li, and Lin}]{zhu2021cylindrical}
Zhu, X.; Zhou, H.; Wang, T.; Hong, F.; Ma, Y.; Li, W.; Li, H.; and Lin, D. 2021.
\newblock Cylindrical and asymmetrical 3d convolution networks for lidar segmentation.
\newblock In \emph{Proceedings of the IEEE/CVF conference on computer vision and pattern recognition}, 9939--9948.

\end{thebibliography}

\end{document}